%% file: RealLFSR_ready.tex
\documentclass[10pt,twocolumn,letterpaper]{article}
 
\usepackage[pagenumbers]{cvpr} 

\usepackage{graphicx}
\usepackage{amsmath}
\usepackage{amssymb}
\usepackage{booktabs}
\usepackage{multirow} 
\usepackage{arydshln}
\usepackage{color}
\usepackage{bm}
\usepackage{colortbl}  
\usepackage{color}
\usepackage{bm}
\usepackage{bbding}
\usepackage{pifont}
\usepackage{wasysym}
\input{includesAndMacros_yagiz}

\usepackage[pagebackref,breaklinks,colorlinks]{hyperref}
 
\definecolor{citecolor}{HTML}{229954}
\definecolor{best}{RGB}{225, 225, 225}

\usepackage[capitalize]{cleveref}
\crefname{section}{Sec.}{Secs.}
\Crefname{section}{Section}{Sections}
\Crefname{table}{Table}{Tables}
\crefname{table}{Tab.}{Tabs.}

\def\confName{CVPR}
\def\confYear{2023}

\begin{document}

\title{Toward Real-World Light Field Super-Resolution}

\author{
Zeyu Xiao$^*$  \quad 
Ruisheng Gao$^*$ \quad 
Yutong Liu \quad 
Yueyi Zhang \quad 
Zhiwei Xiong$^\dagger$\\
University of Science and Technology of China \\
\tt\small \{zeyuxiao,grsmc4180,ustclyt\}@mail.ustc.edu.cn \quad \{zhyuey,zwxiong\}@ustc.edu.cn
}

\twocolumn[{%
	\renewcommand\twocolumn[1][]{#1}%
	\maketitle
	
\vspace{-2mm}
\begin{center}
	\begin{minipage}{1\linewidth}
			\footnotesize
			\centering			
			\renewcommand{\tabcolsep}{0.5pt} 
			\renewcommand{\arraystretch}{0.5} 

			\begin{tabular}{ccccccccccccccc} 
				 
				\multicolumn{3}{c}{\includegraphics[width=0.1985\linewidth]{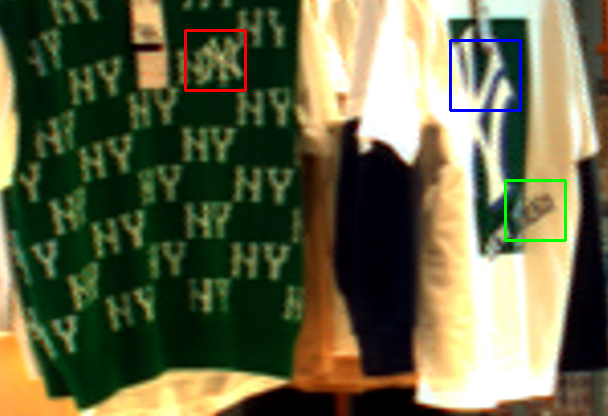}}&
				\multicolumn{3}{c}{\includegraphics[width=0.1985\linewidth]{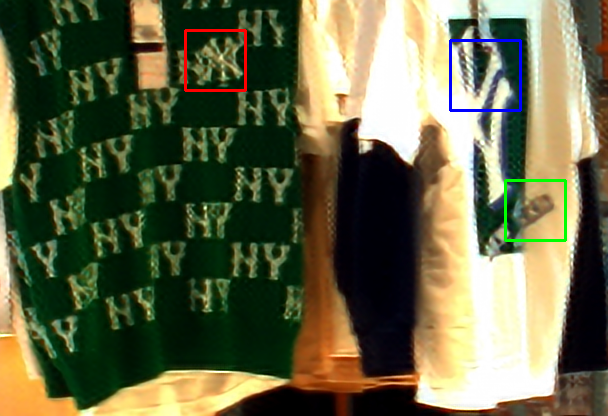}} &
				\multicolumn{3}{c}{\includegraphics[width=0.1985\linewidth]{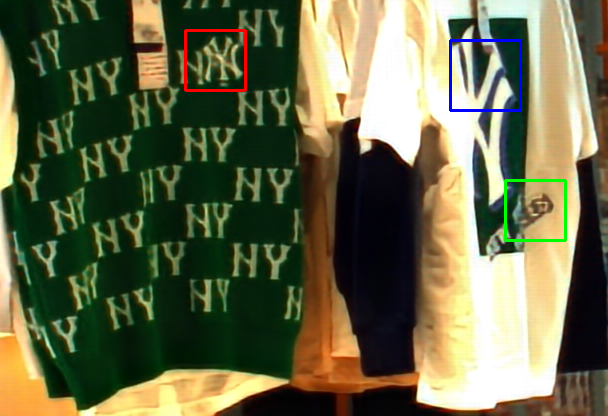}}& 
				\multicolumn{3}{c}{\includegraphics[width=0.1985\linewidth]{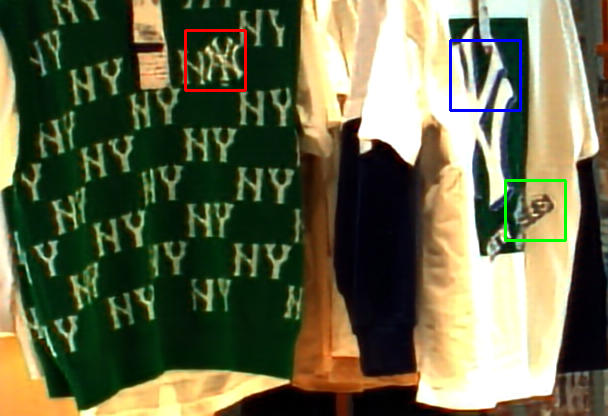}}& 
				\multicolumn{3}{c}{\includegraphics[width=0.1985\linewidth]{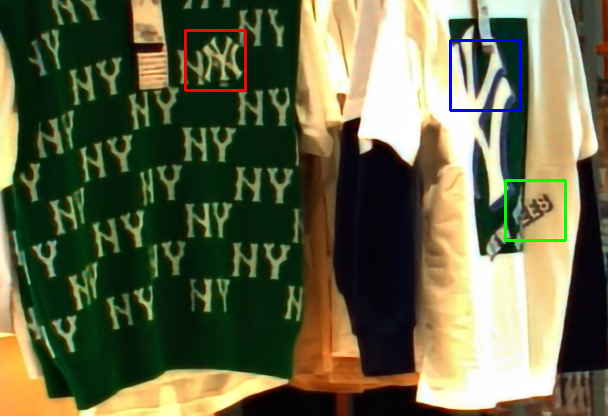}}  \\
				
				\includegraphics[width=0.0645\linewidth]{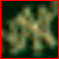} &
				\includegraphics[width=0.0645\linewidth]{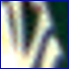} &
				\includegraphics[width=0.0645\linewidth]{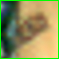} &
				
				\includegraphics[width=0.0645\linewidth]{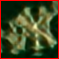} &
				\includegraphics[width=0.0645\linewidth]{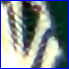} &
				\includegraphics[width=0.0645\linewidth]{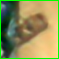} &
				
				\includegraphics[width=0.0645\linewidth]{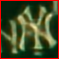} &
				\includegraphics[width=0.0645\linewidth]{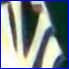} &
				\includegraphics[width=0.0645\linewidth]{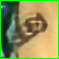} &
				
				\includegraphics[width=0.0645\linewidth]{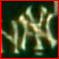} &
				\includegraphics[width=0.0645\linewidth]{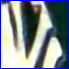} &
				\includegraphics[width=0.0645\linewidth]{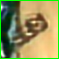} &
				  
				\includegraphics[width=0.0645\linewidth]{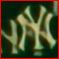} &
				\includegraphics[width=0.0645\linewidth]{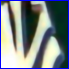} &
				\includegraphics[width=0.0645\linewidth]{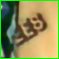} \vspace{1mm}\\

	\multicolumn{3}{c}{(a) Input} &
	\multicolumn{3}{c}{(b) IINet + BI} &
	\multicolumn{3}{c}{(c) IINet + LytroZoom-P} &
	\multicolumn{3}{c}{(c) IINet + LytroZoom-O} &
	\multicolumn{3}{c}{(d) OFPNet + LytroZoom-O}  \\		
				
			\end{tabular} 
\end{minipage}
			\vspace{-3mm}
			\captionof{figure}{
				Examples of $\times 4$ real-world light field SR.
                We collect the first real-world paired LR-HR light field SR dataset (LytroZoom) with authentic degradations to address the real-world light field SR task.
                We show the super-resolved results of existing advanced light field SR method and our proposed OFPNet.
                We show (a) a real-world light field central view image captured by a Lytro ILLUM camera at a focal length of 60 mm,
                SR results generated by (b) IINet~\cite{liu2021intra} trained on the mixed bicubic downsampled synthetic dataset,
                (c) IINet trained on LytroZoom-P, (d) IINet fine-tuned on LytroZoom-O, and (e) the proposed OFPNet fine-tuned on LytroZoom-O.
      }
			\label{fig:teaser} 
\end{center}
}]

\begin{abstract} 

\noindent{\let\thefootnote\relax\footnote{{$^*$ Both authors contribute equally to this work.}}}
\noindent{\let\thefootnote\relax\footnote{{$^\dagger$ Corresponding author.}}}
Deep learning has opened up new possibilities for light field super-resolution (SR),
but existing methods trained on synthetic datasets with simple degradations (e.g., bicubic downsampling) suffer from poor performance when applied to complex real-world scenarios.
To address this problem, we introduce LytroZoom, the first real-world light field SR dataset capturing paired low- and high-resolution light fields of diverse indoor and outdoor scenes using a Lytro ILLUM camera. 
Additionally, we propose the \underline{O}mni-\underline{F}requency \underline{P}rojection \underline{Net}work (OFPNet), which decomposes the omni-frequency components and iteratively enhances them through frequency projection operations to address spatially variant degradation processes present in all frequency components. 
Experiments demonstrate that models trained on LytroZoom outperform those trained on synthetic datasets and are generalizable to diverse content and devices.
Quantitative and qualitative evaluations verify the superiority of OFPNet.
We believe this work will inspire future research in real-world light field SR.
Code and dataset are available at \url{https://github.com/zeyuxiao1997/RealLFSR}.

\end{abstract}

\vspace{-5mm}
\section{Introduction} 
\label{section:inro}

The light field imaging technique enables the capture of the light rays not only at different locations but also from different directions~\cite{bergen1991plenoptic}.
The limited spatial resolution caused by the essential spatial-angular trade-off restricts the capability of light field in practical applications, such as post-capture refocusing~\cite{ng2005light,wang2018selective}, disparity estimation~\cite{wang2015occlusion,zhang2016robust,wang2022disentangling}, and seeing through occlusions~\cite{joshi2007synthetic,wang2020deoccnet,zhang2021removing}.
Attempting to recover a high-resolution (HR) light field from its low-resolution (LR) observation, light field super-resolution (SR) has emerged as a significant task in both academia and industry.
Benefitting from the rapid development of deep learning techniques, convolutional neural network (CNN) based and vision Transformers based methods have demonstrated promising performance for light field SR~\cite{yoon2017light,yeung2018light,wang2018lfnet,gul2018spatial,yuan2018light,meng2019high,cheng2019light,zhang2019residual,wang2020light,jin2020light,wang2020spatial,liu2021intra,zhang2021end,wang2022disentangling,liang2022light,wang2022detail,liang2023learning,xiao2023cutmib}.
They outperform traditional non-learning-based methods~\cite{rossi2017graph,alain2018light,cheng2019light} with appreciable improvements.

However, existing deep methods remain limited because they are trained on simulated light field datasets which assume simple and uniform degradations (\textit{e.g.}, bicubic downsampling) due to the natural difficulty of collecting LR-HR light field pairs.
Degradations in real applications are much more complicated and such simulated degradations usually deviate from real ones. 
This degradation mismatch makes existing light field SR methods unpractical in real-world scenarios~\cite{cheng2019light2,cheng2021light}.
Fig.~\ref{fig:teaser} shows the light field SR results of a real-world light field captured by a Lytro ILLUM camera.
We utilize the advanced IINet~\cite{wang2022disentangling} to train several light field SR models using the mixed simulated dataset with bicubic degradation (IINet + BI) and light field pairs with authentic distortions in our LytroZoom (IINet + LytroZoom-P/LytroZoom-O).
The results clearly show that, the IINet trained on the simulated dataset (Fig.~\ref{fig:teaser}(b)) is less effective in super-resolving on a real-world light field (\textit{e.g.}, a light field captured by Lytro ILLUM).

\begin{figure}[!t]
	\centering
	\includegraphics[width=\linewidth]{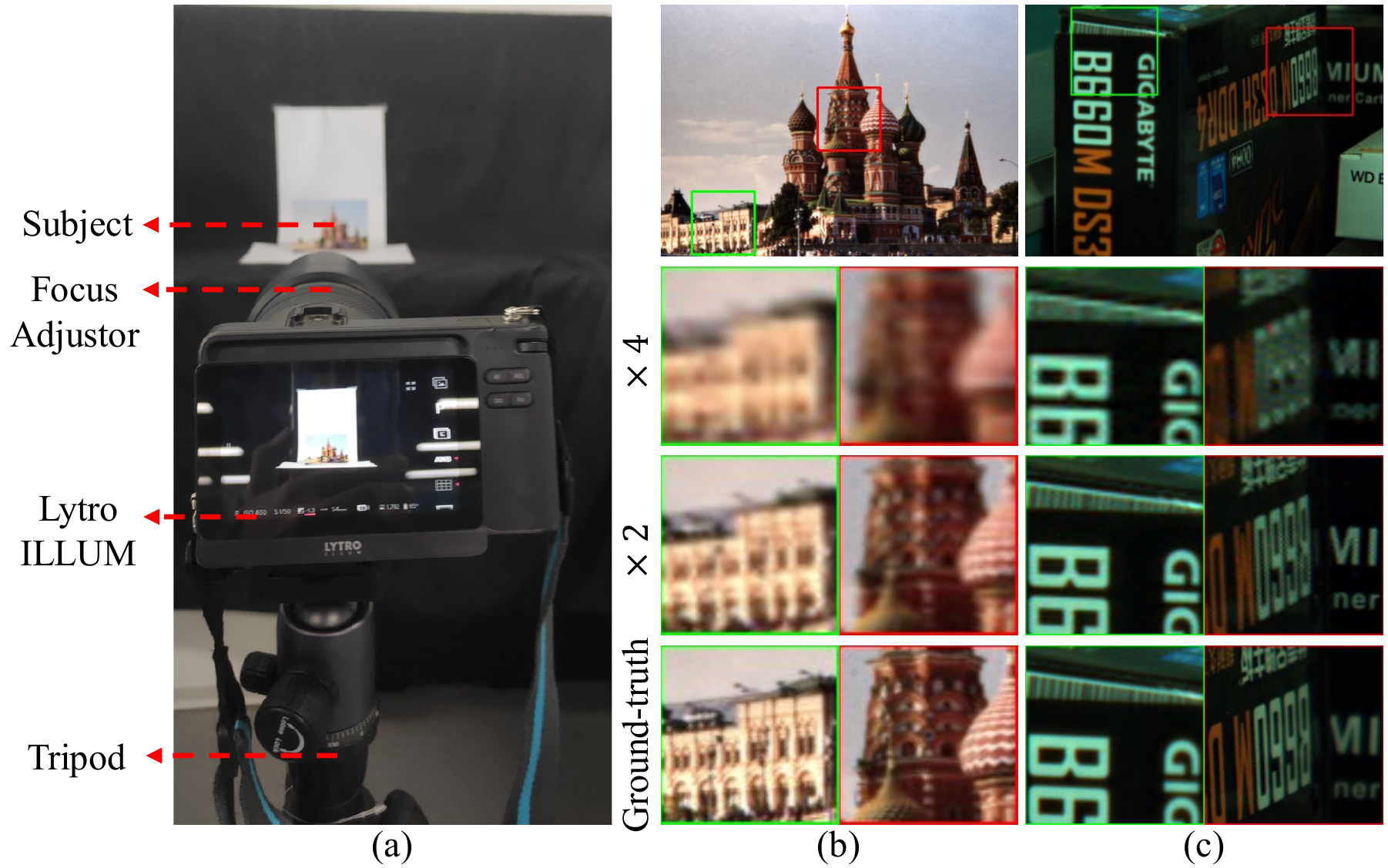}
	\vspace{-7mm}
	\caption{
		(a) The capturing system of the LytroZoom dataset.
		(b) One example of pixel-wise aligned pairs from LytroZoom-P.
		(c) One example of pixel-wise aligned pairs from LytroZoom-O.
	  }
	  \vspace{-5mm}
	\label{fig:lytro}
  \end{figure}

To remedy the above mentioned problem, it is highly desired that we have a light field SR dataset of paired LR-HR pairs more consistent with real-world degradations.
However, collecting such a real-world light field SR dataset is non-trivial since the ground-truth HR light fields are tough to obtain.
Inspired by \cite{chen2019camera}, in which a real-world single image SR dataset is built upon the intrinsic resolution and field-of-view (FoV) degradation in realistic imaging systems,
we capture images of the same scene (\textit{i.e.}, indoor and outdoor scenes) using a Lytro ILLUM camera (Fig.~\ref{fig:lytro}(a)) with different adjusted focal lengths. 
LR and HR light field pairs at different scales (\textit{e.g.}, $\times 2$ and $\times 4$) can be collected by adjusting the focal length.
We utilize the alignment algorithm proposed in \cite{cai2019toward} to rectify distortions caused by different FoVs, including spatial misalignment, intensity variation, and color mismatching.
Scenes that could not be rectified are eliminated from the dataset.
As a result, we collect LytroZoom, a dataset of 94 aligned pairs featuring city scenes printed on postcards (LytroZoom-P), as well as 63 aligned pairs captured with outdoor scenes (LytroZoom-O), as illustrated in Fig.~\ref{fig:lytro}(b) and Fig.~\ref{fig:lytro}(c).
Featuring scenes with spatial diversity and fine-grained details printed on postcards, and depth variations in outdoor scenes, LytroZoom presents a benchmark dataset for light field SR algorithms in real-world scenarios.
As can be seen in Fig.~\ref{fig:teaser}(c)-(d), IINet trained on LytroZoom-P and fine-tuned on LytroZoom-O delivers much better results than the one trained on the simulated data.

Compared with those simulated datasets, the degradation process in LytroZoom is much more complicated. 
In particular, real-world degradations exists in all frequency components and are spatially variant~\cite{fritsche2019frequency,pang2020fan,fuoli2021fourier}.
This motivates us to design a network which can consider omni-frequency information and enhance cross-frequency representations. 
In this paper, we propose the \underline{O}mni-\underline{F}requency \underline{P}rojection \underline{Net}work (OFPNet) to efficiently solve the real-world light field SR problem.
Specifically, we first decompose the input LR light field into high-frequency, middle-frequency and low-frequency components.
Then, in order to improve the corresponding frequency representations, we employ multiple interactive branches that include frequency projection (FP) operations.
Cores of the FP operation are iterative upsampling and downsampling layers to learn non-linear relationships between LR and HR frequency components for super-resolving a real-world light field.
As shown in Fig.~\ref{fig:teaser}(e), our OFPNet can generate better results.

This work makes three key contributions.

(1) We collect LytroZoom, the first real-world paired LR-HR light field SR dataset, which overcomes the limitations of synthetic light field SR datasets and provides a new benchmark for training and evaluating real-world light field SR methods.

(2) We demonstrate that light field SR models trained on LytroZoom perform better on real-world light fields than those trained on synthetic datasets.

(3) We propose OFPNet, a novel baseline network for real-world light field SR, achieving superior results compared to existing methods.

\begin{figure*}[!t]
	\centering
	\includegraphics[width=0.99\textwidth]{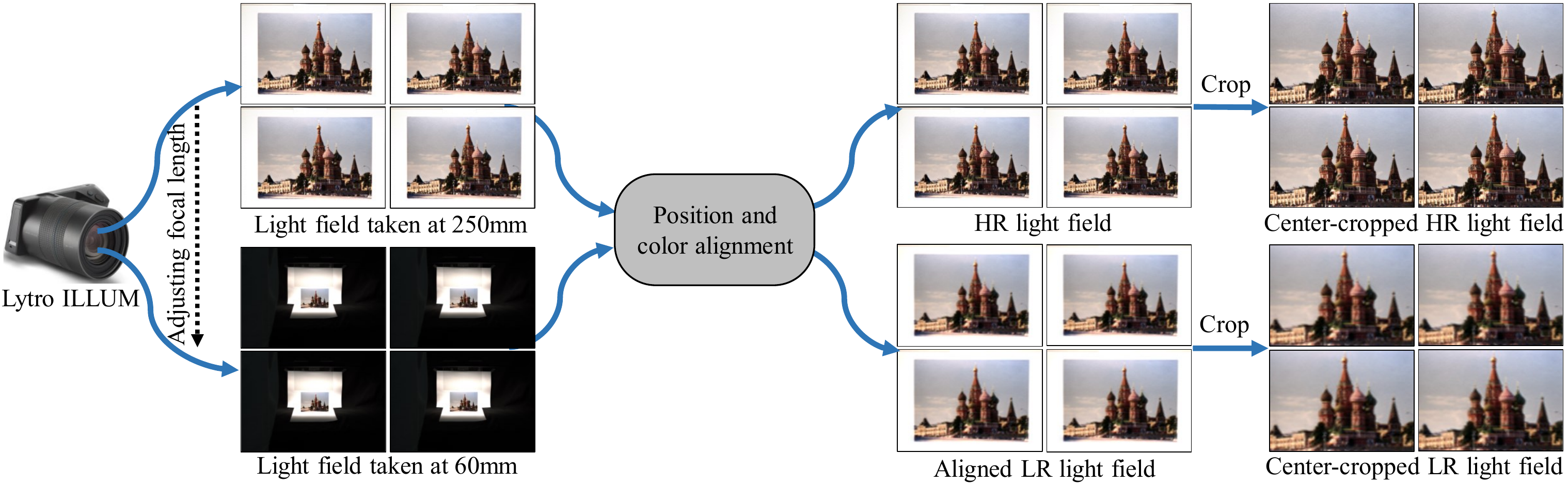}
	\vspace{-3mm}
	\caption{
		The processing pipeline of LytroZoom.
		Here we show how to obtain a $\times 4$ LR-HR real-world light field pair from the postcard.
	  }
	  \vspace{-4mm}
	\label{fig:processpipeline}
\end{figure*}   

\vspace{-1mm}
\section{Related work} 
\vspace{-1mm}

\noindent\textbf{Single image SR.}
Single image SR methods have traditionally been trained on manually synthesized LR images using pre-defined downsampling kernels, such as bicubic interpolation~\cite{dong2014learning,kim2016deeply,lim2017enhanced,tai2017image,zhang2018residual,zhang2018image,li2018multi,dai2019second,hui2019lightweight,liu2020residual,niu2020single,li2020mdcn,gao2022feature,mei2021image,liang2021swinir}.
However, these methods are not directly applicable to real-world images due to their more complex degradation kernels.
Blind SR methods have recently emerged as a solution, which can be categorized into explicit degradation modeling~\cite{bell2019blind,zhang2018learning,gu2019blind,huang2020unfolding,wang2021unsupervised,kim2021koalanet}, capture training pairs~\cite{zhang2019zoom,chen2019camera,cai2019toward,wei2020component,wang2021dual} and generate training pairs~\cite{yuan2018unsupervised,ji2020real,lugmayr2019unsupervised,zhou2019kernel}.
Inspired by the success of single image SR, we propose LytroZoom, the first real-world paired LR-HR light field SR dataset, and a novel OFPNet for real-world light field SR.
The view-dependent degradations in LytroZoom are implicitly modeled by the OFPNet, providing an effective solution for real-world light field SR.

\noindent\textbf{Light field SR.}
Traditional non-learning methods depend on geometric~\cite{liang2015light,rossi2017graph} and mathematical~\cite{alain2018light} modeling of the 4D light field structure to super-resolve the reference view through projection and optimization techniques.
Deep methods now dominate light field SR due to their promising performance.
As a pioneering work along this line, Yoon~\textit{et al.}~\cite{yoon2017light} propose the first light field SR network LFCNN by reusing the SRCNN~\cite{dong2014learning} architecture with multiple channels.
After that, several methods have been designed to exploit across-view redundancy in the light field, either explicitly~\cite{cheng2019light,jin2020light,wang2018lfnet,zhang2019residual} or implicitly~\cite{meng2019high,wang2020spatial,yeung2018light,yuan2018light,wang2022disentangling}.
Transformer-based methods have recently demonstrated the effectiveness in light field SR~\cite{liang2022light,wang2022detail,liang2023learning}.
Recently, Cheng~\textit{et al.}~\cite{cheng2021light} propose a zero-shot learning framework to solve the domain shift problem in light field SR methods.
This work can be regarded as a step towards light field SR in real-world scenarios, but real degradations in light field SR are not approached.
In this paper, we collect the first real-world paired LR-HR light field SR dataset.

\noindent\textbf{Light field SR datasets.}
Several popular datasets, including EPFL~\cite{EPFL}, HCInew~\cite{HCInew}, HCIold~\cite{HCIold}, INRIA~\cite{INRIA}, STFgantry~\cite{StfGantry}, and STFlytro~\cite{StfLytro} are widely used for training and evaluating light field SR methods.
Recently, Sheng~\textit{et al.}~\cite{sheng2022urbanlf} collect UrbanLF, a comprehensive light field dataset containing complex urban scenes for the task of light field semantic segmentation. This dataset has the potential to be extended to light field SR as well.
In all these datasets, the LR light fields are synthesized by the bicubic downsampling operation.
The light field SR models trained on the simulated pairs may exhibit poor performance when applied to real LR light fields where the degradations deviate from the simulated ones~\cite{wang2022learning}.
In this paper, we collect the first real-world paired LR-HR light field SR dataset using a Lytro ILLUM camera.
We capture light fields at multiple focal lengths, providing a general benchmark for real-world light field SR.

\begin{figure*}[!t]
	\centering
	\includegraphics[width=\textwidth]{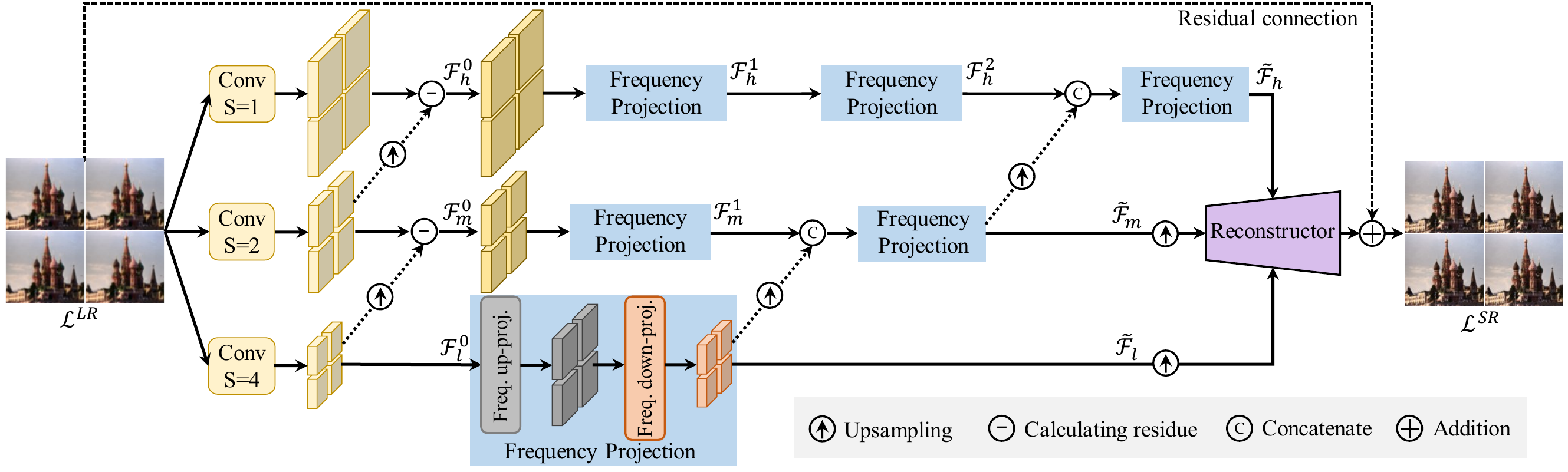}
	\vspace{-7mm}
	\caption{
		Architecture of our proposed OFPNet, which consists of three main components:
		(a) the omni-frequency decomposition which employs three convolutional layers to decompose the input real LR light field $\mathcal{L}^{LR}$ in frequency domain,
		(b) the frequency projection to enhance texture details across different frequency components.
		(c) a reconstrutor aims at aggregating the omni-frequency components for the HR light field reconstruction.
		The reconstructed $\mathcal{L}^{SR}$ is obtained by adding the predicted residual to an LR light field.
	  }
	  \vspace{-3mm}
	\label{fig:pipeline}
  \end{figure*}

\vspace{-1mm}
\section{Capturing the LytroZoom Dataset} 
\vspace{-1mm}
\label{dataset}   

Our goal is to collect a real-world light field SR dataset consisting of paired LR-HR light fields.
This is challenging, as it requires accurately aligned LR-HR sequences of the same scene.
To address this challenge, we use the Lytro ILLUM camera, which can directly display the actual focal lengths during shooting.
We capture city scenes printed on postcards and outdoor static objects as subjects.

We define a light field captured at 250 mm focal length as the HR ground-truth and the ones captured at 120mm and 60mm focal lengths as the $\times 2$ and $\times 4$ LR observations~\cite{chen2019camera,cai2019toward}.
We standardize our data collection process by shooting postcards at the lowest ISO setting and maintaining a constant white balance and aperture size. For outdoor scenes, we adjust the ISO value to ensure optimal exposure and minimize noise.
The camera is fixed on a tripod for stabilization.
Without post-processing procedures, such as color correction and histogram equalization, we decode the captured raw light field data using the light field toolbox~\cite{dansereau2013decoding,dansereau2015linear}.
This results in $15 \times 15 \times 625 \times 434$ light field pairs.
However, spatial misalignment, intensity variation, and color mismatching may exist in different views due to uncontrollable changes during lens zooming.
Therefore, we use the alignment algorithm~\cite{cai2019toward} view-by-view to rectify LR light fields iteratively to preserve view-dependent degradations.
We then center-crop light field pairs to mitigate lens distortion and the vignetting effect.
Please refer to Fig.~\ref{fig:processpipeline} for the entire pipeline.

After careful shooting, decoding, rectifying, cropping, and selection, we collect a dataset with 94 city scenes printed on postcards and 63 outdoor static scenes.
We provide pixel-wise aligned light field pairs (\textit{i.e.}, ground-truth, $\times 2$ and $\times 4$ LR observations) with the resolution of $5 \times 5 \times 456 \times 320$ (LytroZoom-P) and $5 \times 5 \times 608 \times 416$ (LytroZoom-O).
Fig.~\ref{fig:lytro} shows samples from the dataset.
We randomly partition LytroZoom-P into 63 scenes for training, 17 for validation, and the remaining 15 scenes for testing. Similarly, we use 55 scenes for training and the remaining 10 scenes for testing in LytroZoom-O.

\vspace{-1mm}
\section{Omni-Frequency Projection Network} 
\vspace{-1mm}

In Sec.~\ref{dataset}, we have collected LytroZoom, containing diverse contents captured by a Lytro ILLUM camera.
Therefore we have access to the LR-HR pairs for training light field SR networks.
However, existing light field SR methods do not entirely account for the characteristics of real-degraded light fields.
To address this, we propose OFPNet for real-world light field SR, which can model real degradations in real-world light field SR.
 
\subsection{Overview}
\label{overview}
Following~\cite{yuan2018light,wang2018lfnet,yeung2018light,zhang2019residual,wang2020spatial,wang2020light}, we super-resolve the Y channel images, leaving Cb and Cr channel images being bicubic upscaled for light field SR.
Without considering the channel dimension and given an LR light field $\mathcal{L}^{LR} \in \mathbb{R}^{U \times V \times H \times W}$ with few details and textures, we aim at generating an HR light field $\mathcal{L}^{SR} \in \mathbb{R}^{U \times V \times H \times W}$ with more details, which should be close to the ground-truth $\mathcal{L}^{GT} \in \mathbb{R}^{U \times V \times H \times W}$.
$U$ and $V$ represent angular dimensions, and $H$ and $W$ represent spatial dimensions.

Inspired by recent progress in real-world single image SR, the degradation exists in all frequency components~\cite{fritsche2019frequency,zhou2020guided,ji2021frequency,pang2020fan,li2021learning}, we propose OFPNet, in which omni-frequency components are considered for real-world light field SR.
In OFPNet, we first decompose the input LR light field $\mathcal{L}^{LR}$ into high-frequency, middle-frequency, and low-frequency components, \textit{i.e.}, $\mathcal{F}_h$, $\mathcal{F}_m$, and $\mathcal{F}_l$.
Then the FP operations are utilized on three frequency branches to enhance frequency representations.
These branches are interacted to enhance the cross-frequency representations.
After we obtain the enhanced frequency features $\tilde{\mathcal{F}}_h$, $\tilde{\mathcal{F}}_m$, and $\tilde{\mathcal{F}}_l$, we feed them to the reconstructor to generate $\mathcal{L}^{SR}$.

\subsection{Omni-Frequency Decomposition} 
\label{decomposition}
To obtain the informative omni-frequency representation for real-world light field SR, we first decompose $\mathcal{L}^{LR}$ into different frequency components.
We utilize the learnable spatial downsampling operations to decompose frequency components in the feature domain, which is in a spirit similar to the octave convolution~\cite{chen2019drop,akbari2020generalized}.

As shown in Fig.~\ref{fig:pipeline}, we first downsample $\mathcal{L}^{LR}$ by a convolution layer with $\text{stride}\!=\!4$ to get the corresponding low-frequency component $\mathcal{F}_l$.
Then we obtain the middle frequency component $\mathcal{F}_m$ by removing $\mathcal{F}_l$ from the corresponding original feature, which is downsampled with a $\text{stride}\!=\!2$ convolutional layer. 
Similarly, to get the high frequency component $\mathcal{F}_h$, we remove the downsampled feature with $\text{stride}\!=\!2$ from the feature extracted with $\text{stride}\!=\!1$ convolutional layer.
The whole process can be denoted as
\begin{equation}
	\begin{array}{l}
	\mathcal{F}_{l}=\text{conv}_{2}\left(\text{conv}_{2}(\mathcal{L}^{LR})\right), \\
	\mathcal{F}_{m}=\text{conv}_{2}(\mathcal{L}^{LR})-[\text{conv}_{2}\left(\text{conv}_{2}(\mathcal{L}^{LR})\right)] \uparrow_{2}, \\
	\mathcal{F}_{h}=\text{conv}(\mathcal{L}^{LR})-[\text{conv}_{2}(\mathcal{L}^{LR})] \uparrow_{2},
	\end{array}
\end{equation}
where $\text{conv}(\cdot)$ denotes the convolution layer without downsampling and $\text{conv}_{2}(\cdot)$ denotes the convolution layer with $\text{stride}\!=\!2$.
$[\cdot]\uparrow_{r}$ means the bilinear upsampling operation with the $\text{factor}\!=\!r$.

\subsection{Frequency Projection}
The extracted frequency components face the inevitable information loss problem caused by irreversible convolutional layers.
Inspired by the back-projection operation that produces an HR feature map and iteratively refines it through multiple upsampling and downsampling layers to learn nonlinear relationships between LR and HR images~\cite{haris2018deep,haris2019recurrent,haris2020deep,hu2022spatial}, we introduce the FP operation to enhance frequency feature representations, making up for the information lost.

As shown in Fig.~\ref{fig:pipeline}, the FP operation consists of the frequency up-projection unit (FUPU) and the frequency down-projection unit (FDPU), in which nonlinear relationships between LR and HR frequency features can be exploited iteratively.
We first project the extracted frequency feature $\mathcal{F}^{n-1}$ to corresponding HR representation $U^{n-1}$ based on a frequency scale-up block 
\begin{equation}
	{U}^{n-1} = \text{Up}(\mathcal{F}^{n-1}),
\end{equation} 
where $\text{Up}(\cdot)$ denotes the frequency scale-up block.
It first fuses multi-view information progressively using the residual blocks, in which the inter-view correlations can be exploited, then upsamples the fused feature by bilinear interpolation followed by a $1 \times 1$ convolutional layer~\cite{liu2021intra}.
Please refer to the supplementary document for the detailed structure.
$m$ denotes the number of FUPU.

Then we project the HR representation back to LR one and obtain the corresponding residuals $e^{n-1}$ between the back-projected representation and original LR input
\begin{equation}
	e^{n-1} = \text{Down}({U}^{n-1})-\mathcal{F}^{n-1},
\end{equation}
where $\text{Down}(\cdot)$ denotes the frequency scale-down block.
It first reduces the resolution of ${U}^{n-1}$ to the original one via a $4 \times 4$ convolutional layer with $\text{stride}\!=\!2$, followed by fusing multi-view information progressively.

Finally, we back-project the residual to the HR representation and eliminate the corresponding super-resolved representation errors to obtain the final output of FUPU
\begin{equation}
	U^n = \text{Up}(e^{n-1})+{U}^{n-1}.
\end{equation}

The procedure for FDPU is similar to FUPU.
FDPU aims to obtain refined LR frequency representations by projecting the previously updated HR frequency representation.
Please see the supplementary document for more details.

We can enhance the representations of different frequency components thanks to the FP operations.
In practice, however, the high-frequency component is relatively tricky to enhance~\cite{fritsche2019frequency,wei2020component}.
We, therefore, propose to enhance such challenging frequency components in a coarse-to-fine manner.
Specifically, we encourage the interaction between different frequency components and progressively utilize the enhanced lower frequency representations to help the enhancement of higher frequency components by concatenating them together.
The final enhanced frequency features can be denoted as
\begin{equation}
    \begin{array}{l}
        \tilde{\mathcal{F}}_l=\text {FP}_1(\mathcal{F}_h), \\
        \tilde{\mathcal{F}}_m=\text {FP}_2(\text{conv}([\text {FP}_1(\mathcal{F}_m),\tilde{\mathcal{F}}_l])), \\
        \tilde{\mathcal{F}}_h=\text {FP}_3(\text{conv}([\text {FP}_2(\text {FP}_1(\mathcal{F}_h)),\tilde{\mathcal{F}}_m])),
    \end{array}
\end{equation}
where $\text{FP}_n(\cdot)$ denotes the $n$-th FP operation, and $\text{conv}(\cdot)$ here aims at reducing the channel dimensions.

\vspace{-1mm}
\subsection{Reconstructor}
\vspace{-1mm}
We feed the enhanced frequency features to the reconstructor to generate the super-resolved results.
We first concatenate $\tilde{\mathcal{F}}_l$, $\tilde{\mathcal{F}}_m$, $\tilde{\mathcal{F}}_h$ along the channel dimension, followed by a convolutional layer to reduce the channel number.
The concatenated feature is further fed to the Feature Blending Module (FBM) and the Upsampling Module~\cite{liu2021intra}.
Note that we remove the pixel shuffling operation because the LR-HR pairs have the same spatial resolution.
 
The $L_1$-norm loss function is employed to minimize the pixel-wise distance between the generated HR light field $\mathcal{L}^{SR}$ and the ground-truth $\mathcal{L}^{GT}$.
\begin{equation}
    L(\mathcal{L}^{GT}, \mathcal{L}^{SR})=\|\mathcal{L}^{GT}- \mathcal{L}^{SR}\|.
\end{equation}

\begin{figure}[!t]
	\centering
	\includegraphics[width=\linewidth]{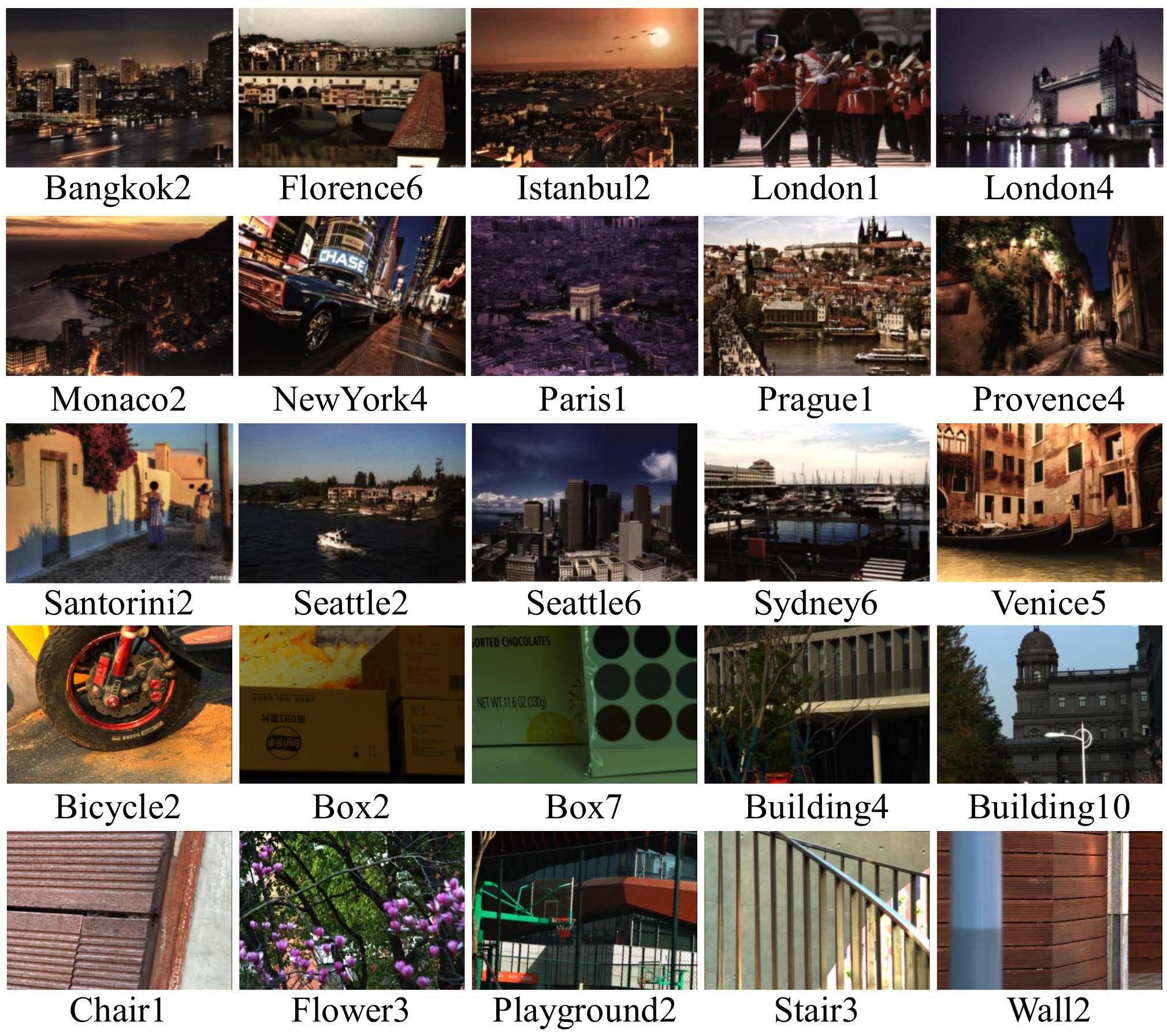}
	\vspace{-7mm}
	\caption{
		Thumbnails of 15 test scenes from LytroZoom-P (first three rows) and 10 test scenes from LytroZoom-O (last two rows).
		We show the ground-truth central view images here. 
	  }
	  \vspace{-3mm} 
	\label{fig:thumbnail}
\end{figure}

\begin{figure*}[!t]
	\centering
        \includegraphics[width=0.99\linewidth]{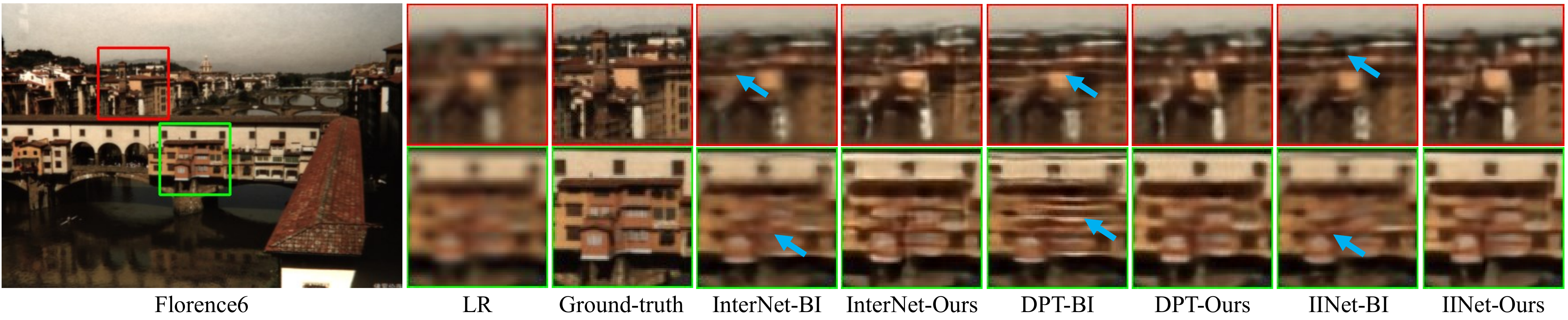}
	\vspace{-3mm} 
	\caption{
		Visual comparisons ($\times 4$ SR) of different models (trained on the BI and LytroZoom-P datasets) on the LytroZoom-P testset.
	  }
	  \vspace{-3mm}
	\label{fig:bi_lytro}
\end{figure*}

\begin{table*}[!t]
	\centering
	\begin{minipage}{0.58\textwidth}
		\caption{
			Average PSNR (dB) and SSIM results on the LytroZoom-P testset by different methods.
			“BI” indicates the method is trained on bicubic-downsampled datasets, and “LZ-P” indicates the method is trained on LytroZoom-P.
		  } 
		\vspace{-3mm}
        \renewcommand\arraystretch{1.05}
	  \resizebox{\textwidth}{!}{
		\scriptsize
		\centering
		\begin{tabular}{|c|c|cc|cc|cc|c|}
			\hline
			\multirow{2}{*}{Metric} & \multirow{2}{*}{Scale} & \multicolumn{2}{c|}{InterNet}        & \multicolumn{2}{c|}{DPT}             & \multicolumn{2}{c|}{IINet}           & \multirow{2}{*}{OFPNet} \\ \cline{3-8}
									&                        & \multicolumn{1}{c|}{BI}     & LZ-P  & \multicolumn{1}{c|}{BI}     & LZ-P  & \multicolumn{1}{c|}{BI}     & LZ-P   &                         \\ 
									\hline
									\hline
			\multirow{2}{*}{PSNR}   & $\times 2$                & \multicolumn{1}{c|}{31.55}  & 38.78  & \multicolumn{1}{c|}{31.55}  & 38.65  & \multicolumn{1}{c|}{31.88}  & 38.78  & \textbf{38.89}             \\ \cline{2-9} 
									& $\times 4$                & \multicolumn{1}{c|}{26.12}       & 29.60  & \multicolumn{1}{c|}{25.63}       &  29.43  & \multicolumn{1}{c|}{26.04}       & 29.82  & \textbf{30.11}                   \\ 
									\hline
									\hline
			\multirow{2}{*}{SSIM}   & $\times 2$                      & \multicolumn{1}{c|}{0.9138} & 0.9764 & \multicolumn{1}{c|}{0.9145} & 0.9754 & \multicolumn{1}{c|}{0.9170} & 0.9771 &  \textbf{0.9779}            \\ \cline{2-9} 
									& $\times 4$                      & \multicolumn{1}{c|}{0.7685}       &   0.8626  & \multicolumn{1}{c|}{0.7569}       &   0.8560  & \multicolumn{1}{c|}{0.7621}       &   0.8706 &  \textbf{0.8786}                   \\ 
									\hline
			\end{tabular}
	  } 
	  \label{tab:table}
	\end{minipage}
	\hspace{3mm}
	\begin{minipage}{0.39\textwidth}
	  \caption{
	Average PSNR (dB) and SSIM results on LytroZoom-O testset by different methods (pretrained on LytroZoom-P and fine-tuned on LytroZoom-O).
  	} 
	  \vspace{-3mm}
        \renewcommand\arraystretch{1.05}
	  \resizebox{\textwidth}{!}{
		\scriptsize
		\centering
		\begin{tabular}{|c|c|c|c|c|c|}
			\hline
			{Metric} & {Scale} & {InterNet} & {DPT}   & {IINet}    & {OFPNet} \\
									\hline
									\hline
			\multirow{2}{*}{PSNR}   & $\times 2$  & 30.15      & 30.08   & 30.28 & \textbf{30.79}    \\ \cline{2-6} 
									& $\times 4$   & 27.88     &  27.95   & 28.22  & \textbf{28.91}    \\ 
									\hline
									\hline
			\multirow{2}{*}{SSIM}   & $\times 2$   & 0.8738     & 0.8603   & 0.8737 & \textbf{0.8863}    \\ \cline{2-6} 
									& $\times 4$   & 0.7627     &  0.7620   &  0.7770 & \textbf{0.7991}   \\ 
									\hline
			
			\end{tabular}
	  } 
	  \label{tab:table2}
	\end{minipage}
	\vspace{-4mm}
  \end{table*}

\vspace{-3mm}
\section{Experiments}
\vspace{-1mm}
\subsection{Experimental Settings}
\vspace{-1mm}

\noindent \textbf{Inference settings.}
PSNR and SSIM (the higher, the better) are adopted to evaluate the reconstruction accuracy.
Following ~\cite{yuan2018light,wang2018lfnet,yeung2018light,zhang2019residual,wang2020spatial,wang2020light}, the light field SR results are evaluated using PSNR and SSIM indices on the Y channel in the YCbCr space.
We evaluate the performance of different networks under the settings of $\times 2$ and $\times 4$ light field SR.
To compare the angular consistency of the reconstructed HR results, the epipolar plane images (EPIs) are visualized for quantitative comparison in this paper.

\noindent \textbf{Selected baseline methods.}
We select three representative and advanced light field SR networks, \textit{i.e.}, InterNet~\cite{wang2020spatial}, DPT~\cite{wang2022detail}, and IINet~\cite{liu2021intra}, as our main baselines.
Note that, we find that ATO~\cite{jin2020light} and DistgSSR~\cite{wang2022disentangling} cannot converge on LytroZoom, so we do not include them in the comparison. 
We follow the same experimental settings reported in their paper and retrain these networks based on their publicly available codes.
Note that, we do not compare methods that require large memory consumptions during the training stage (\textit{e.g.}, LFT~\cite{liang2022light}).
We also exclude single image SR methods because previous work~\cite{yuan2018light,wang2018lfnet,yeung2018light,zhang2019residual,wang2020spatial,wang2020light} has proved that these methods cannot generate results with high fidelity and good angular consistency.

\noindent \textbf{Implementation and training details of OFPNet.}
In our implementation, the channel number is set to 32 unless otherwise specified.
We utilize the Adam optimizer with parameters $\beta_1 = 0.9$ and $\beta_2 = 0.999$.
Each mini-batch consists of 2 samples with $72 \times 72$ patches for $\times 2$ light field SR and 4 samples with $64 \times 64$ patches for $\times 4$ light field SR.
We first train OFPNet on the LytroZoom-P dataset and then fine-tune the pretrained OFPNet on LytroZoom-O.
The initial learning rate is set to $1e-4$, and we reduce it by a factor of 0.5 every 2,000 epochs until 8,000 epochs during the training stage.
During the fine-tuning stage, the initial learning rate is also set to $1e-4$, and we reduce it by a factor of 0.5 every 1,000 epochs until 5,000 epochs.
OFPNet is trained and fine-tuned on two NVIDIA GTX 1080Ti GPUs.

\begin{figure*}[!t]
	\begin{center}
	\begin{minipage}{0.99\linewidth}
	\begin{minipage}{0.363\linewidth}
	\centerline{\includegraphics[width=1\linewidth]{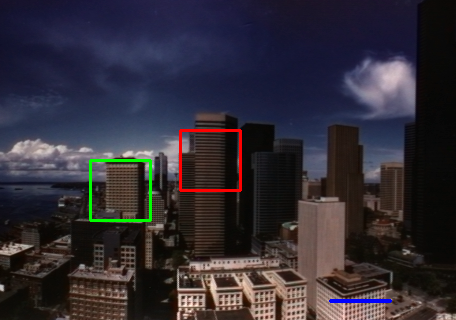}} \vfill \vspace{-0.15cm}
	\centerline{\scriptsize{Seattle6}}
	\end{minipage}
	\hfill
	\hspace{-0.200cm}
	\begin{minipage}{0.10390\linewidth}
	\centerline{\includegraphics[width=1\linewidth]{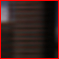}} \vspace{-0.15mm}
	\centerline{\includegraphics[width=1\linewidth]{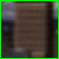}} \vfill \vspace{-0.02cm}
	\centerline{\includegraphics[width=1\linewidth]{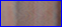}} \vfill \vspace{-0.15cm}
	\centerline{\scriptsize{LR}}
	\end{minipage}
	\hfill
	\hspace{-0.200cm}
	\begin{minipage}{0.10390\linewidth}
	\centerline{\includegraphics[width=1\linewidth]{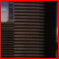}} \vspace{-0.15mm}
	\centerline{\includegraphics[width=1\linewidth]{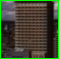}} \vfill \vspace{-0.02cm}
	\centerline{\includegraphics[width=1\linewidth]{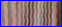}} \vfill \vspace{-0.15cm}
	\centerline{\scriptsize{Ground-truth}}
	\end{minipage}
	\hfill
	\hspace{-0.200cm}
	\begin{minipage}{0.10390\linewidth}
	\centerline{\includegraphics[width=1\linewidth]{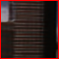}} \vspace{-0.15mm}
	\centerline{\includegraphics[width=1\linewidth]{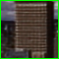}} \vfill \vspace{-0.02cm}
	\centerline{\includegraphics[width=1\linewidth]{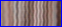}} \vfill \vspace{-0.15cm}
	\centerline{\scriptsize{InterNet}}
	\end{minipage}
	\hfill
	\hspace{-0.200cm}
	\begin{minipage}{0.10390\linewidth}
	\centerline{\includegraphics[width=1\linewidth]{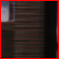}} \vspace{-0.15mm}
	\centerline{\includegraphics[width=1\linewidth]{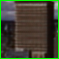}} \vfill \vspace{-0.02cm}
	\centerline{\includegraphics[width=1\linewidth]{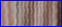}} \vfill \vspace{-0.15cm}
	\centerline{\scriptsize{DPT}}
	\end{minipage}
	\hfill
	\hspace{-0.200cm}
	\begin{minipage}{0.10390\linewidth}
	\centerline{\includegraphics[width=1\linewidth]{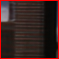}} \vspace{-0.15mm}
	\centerline{\includegraphics[width=1\linewidth]{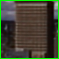}} \vfill \vspace{-0.02cm}
	\centerline{\includegraphics[width=1\linewidth]{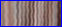}} \vfill \vspace{-0.15cm}
	\centerline{\scriptsize{IINet}}
	\end{minipage}
	\hfill
	\hspace{-0.200cm}
	\begin{minipage}{0.10390\linewidth}
	\centerline{\includegraphics[width=1\linewidth]{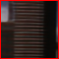}} \vspace{-0.15mm}
	\centerline{\includegraphics[width=1\linewidth]{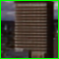}} \vfill \vspace{-0.02cm}
	\centerline{\includegraphics[width=1\linewidth]{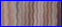}} \vfill \vspace{-0.15cm}
	\centerline{\scriptsize{OFPNet}}
	\end{minipage}
	\vfill
	\vspace{1mm}
	\begin{minipage}{0.363\linewidth}
	\centerline{\includegraphics[width=1\linewidth]{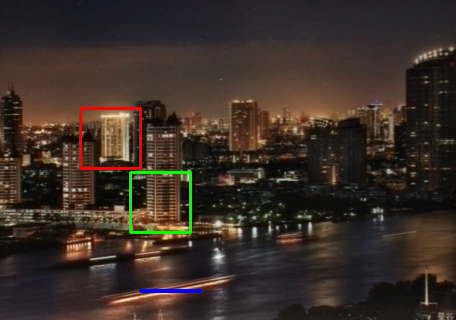}} \vfill \vspace{-0.15cm}
	\centerline{\scriptsize{Bangkok2}}
	\end{minipage}
	\hfill
	\hspace{-0.200cm}
	\begin{minipage}{0.10390\linewidth}
	\centerline{\includegraphics[width=1\linewidth]{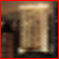}} \vspace{-0.15mm}
	\centerline{\includegraphics[width=1\linewidth]{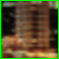}} \vfill \vspace{-0.02cm}
	\centerline{\includegraphics[width=1\linewidth]{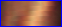}} \vfill \vspace{-0.15cm}
	\centerline{\scriptsize{LR}}
	\end{minipage}
	\hfill
	\hspace{-0.200cm}
	\begin{minipage}{0.10390\linewidth}
	\centerline{\includegraphics[width=1\linewidth]{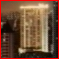}} \vspace{-0.15mm}
	\centerline{\includegraphics[width=1\linewidth]{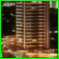}} \vfill \vspace{-0.02cm}
	\centerline{\includegraphics[width=1\linewidth]{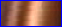}} \vfill \vspace{-0.15cm}
	\centerline{\scriptsize{Ground-truth}}
	\end{minipage}
	\hfill
	\hspace{-0.200cm}
	\begin{minipage}{0.10390\linewidth}
	\centerline{\includegraphics[width=1\linewidth]{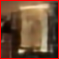}} \vspace{-0.15mm}
	\centerline{\includegraphics[width=1\linewidth]{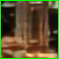}} \vfill \vspace{-0.02cm}
	\centerline{\includegraphics[width=1\linewidth]{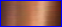}} \vfill \vspace{-0.15cm}
	\centerline{\scriptsize{InterNet}}
	\end{minipage}
	\hfill
	\hspace{-0.200cm}
	\begin{minipage}{0.10390\linewidth}
	\centerline{\includegraphics[width=1\linewidth]{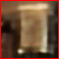}} \vspace{-0.15mm}
	\centerline{\includegraphics[width=1\linewidth]{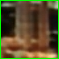}} \vfill \vspace{-0.02cm}
	\centerline{\includegraphics[width=1\linewidth]{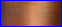}} \vfill \vspace{-0.15cm}
	\centerline{\scriptsize{DPT}}
	\end{minipage}
	\hfill
	\hspace{-0.200cm}
	\begin{minipage}{0.10390\linewidth}
	\centerline{\includegraphics[width=1\linewidth]{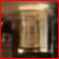}} \vspace{-0.15mm}
	\centerline{\includegraphics[width=1\linewidth]{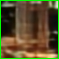}} \vfill \vspace{-0.02cm}
	\centerline{\includegraphics[width=1\linewidth]{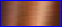}} \vfill \vspace{-0.15cm}
	\centerline{\scriptsize{IINet}}
	\end{minipage}
	\hfill
	\hspace{-0.200cm}
	\begin{minipage}{0.10390\linewidth}
	\centerline{\includegraphics[width=1\linewidth]{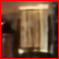}} \vspace{-0.15mm}
	\centerline{\includegraphics[width=1\linewidth]{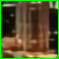}} \vfill \vspace{-0.02cm}
	\centerline{\includegraphics[width=1\linewidth]{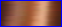}} \vfill \vspace{-0.15cm}
	\centerline{\scriptsize{OFPNet}}
	\end{minipage}
	\end{minipage}
	\end{center}
	\vspace{-7mm}
	\caption{
	  Visual comparisons (central views) of different models (trained on LytroZoom-P) on the LytroZoom-P testset.
	  Top: $\times 2$ SR.
	  Bottom: $\times 4$ SR.
	  Please zoom in for better visualization and best viewed on the screen.
	  }
	\label{fig:compare}
	\vspace{-2mm}
\end{figure*}
  
\begin{figure*}[!t]
	\begin{center}
	\begin{minipage}{0.99\linewidth}
	\begin{minipage}{0.363\linewidth}
	\centerline{\includegraphics[width=1\linewidth]{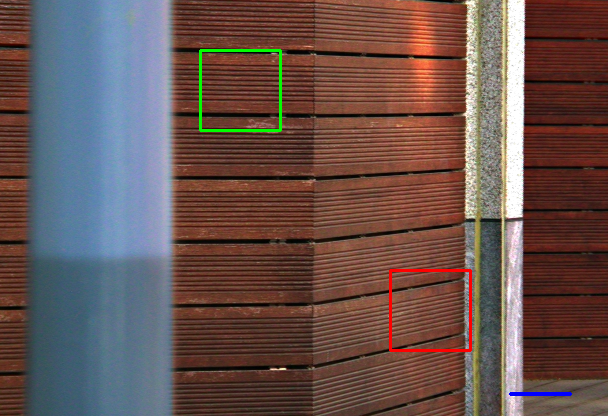}} \vfill \vspace{-0.15cm}
	\centerline{\scriptsize{Wall2}}
	\end{minipage}
	\hfill
	\hspace{-0.200cm}
	\begin{minipage}{0.10390\linewidth}
	\centerline{\includegraphics[width=1\linewidth]{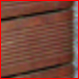}} \vspace{-0.15mm}
	\centerline{\includegraphics[width=1\linewidth]{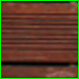}} \vfill \vspace{-0.02cm}
	\centerline{\includegraphics[width=1\linewidth]{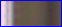}} \vfill \vspace{-0.15cm}
	\centerline{\scriptsize{LR}}
	\end{minipage}
	\hfill
	\hspace{-0.200cm}
	\begin{minipage}{0.10390\linewidth}
	\centerline{\includegraphics[width=1\linewidth]{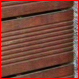}} \vspace{-0.15mm}
	\centerline{\includegraphics[width=1\linewidth]{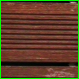}} \vfill \vspace{-0.02cm}
	\centerline{\includegraphics[width=1\linewidth]{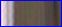}} \vfill \vspace{-0.15cm}
	\centerline{\scriptsize{Ground-truth}}
	\end{minipage}
	\hfill
	\hspace{-0.200cm}
	\begin{minipage}{0.10390\linewidth}
	\centerline{\includegraphics[width=1\linewidth]{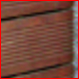}} \vspace{-0.15mm}
	\centerline{\includegraphics[width=1\linewidth]{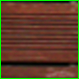}} \vfill \vspace{-0.02cm}
	\centerline{\includegraphics[width=1\linewidth]{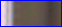}} \vfill \vspace{-0.15cm}
	\centerline{\scriptsize{InterNet}}
	\end{minipage}
	\hfill
	\hspace{-0.200cm}
	\begin{minipage}{0.10390\linewidth}
	\centerline{\includegraphics[width=1\linewidth]{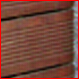}} \vspace{-0.15mm}
	\centerline{\includegraphics[width=1\linewidth]{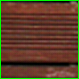}} \vfill \vspace{-0.02cm}
	\centerline{\includegraphics[width=1\linewidth]{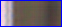}} \vfill \vspace{-0.15cm}
	\centerline{\scriptsize{DPT}}
	\end{minipage}
	\hfill
	\hspace{-0.200cm}
	\begin{minipage}{0.10390\linewidth}
	\centerline{\includegraphics[width=1\linewidth]{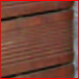}} \vspace{-0.15mm}
	\centerline{\includegraphics[width=1\linewidth]{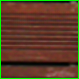}} \vfill \vspace{-0.02cm}
	\centerline{\includegraphics[width=1\linewidth]{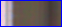}} \vfill \vspace{-0.15cm}
	\centerline{\scriptsize{IINet}}
	\end{minipage}
	\hfill
	\hspace{-0.200cm}
	\begin{minipage}{0.10390\linewidth}
	\centerline{\includegraphics[width=1\linewidth]{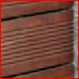}} \vspace{-0.15mm}
	\centerline{\includegraphics[width=1\linewidth]{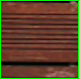}} \vfill \vspace{-0.02cm}
	\centerline{\includegraphics[width=1\linewidth]{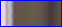}} \vfill \vspace{-0.15cm}
	\centerline{\scriptsize{OFPNet}}
	\end{minipage}
	\vfill
	\vspace{1mm}
	\begin{minipage}{0.363\linewidth}
	\centerline{\includegraphics[width=1\linewidth]{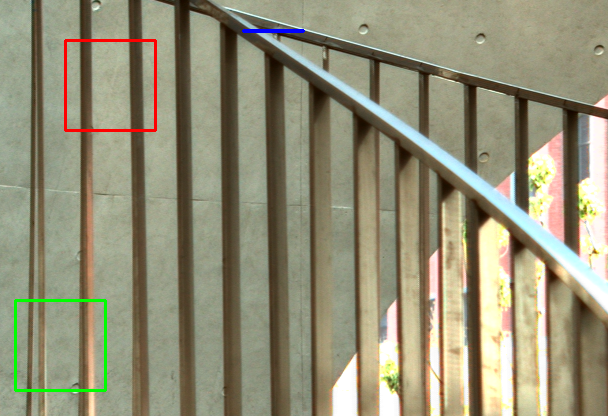}} \vfill \vspace{-0.15cm}
	\centerline{\scriptsize{Stair3}}
	\end{minipage}
	\hfill
	\hspace{-0.200cm}
	\begin{minipage}{0.10390\linewidth}
	\centerline{\includegraphics[width=1\linewidth]{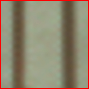}} \vspace{-0.15mm}
	\centerline{\includegraphics[width=1\linewidth]{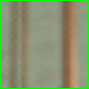}} \vfill \vspace{-0.02cm}
	\centerline{\includegraphics[width=1\linewidth]{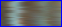}} \vfill \vspace{-0.15cm}
	\centerline{\scriptsize{LR}}
	\end{minipage}
	\hfill
	\hspace{-0.200cm}
	\begin{minipage}{0.10390\linewidth}
	\centerline{\includegraphics[width=1\linewidth]{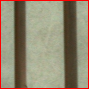}} \vspace{-0.15mm}
	\centerline{\includegraphics[width=1\linewidth]{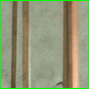}} \vfill \vspace{-0.02cm}
	\centerline{\includegraphics[width=1\linewidth]{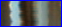}} \vfill \vspace{-0.15cm}
	\centerline{\scriptsize{Ground-truth}}
	\end{minipage}
	\hfill
	\hspace{-0.200cm}
	\begin{minipage}{0.10390\linewidth}
	\centerline{\includegraphics[width=1\linewidth]{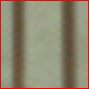}} \vspace{-0.15mm}
	\centerline{\includegraphics[width=1\linewidth]{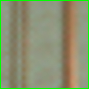}} \vfill \vspace{-0.02cm}
	\centerline{\includegraphics[width=1\linewidth]{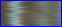}} \vfill \vspace{-0.15cm}
	\centerline{\scriptsize{InterNet}}
	\end{minipage}
	\hfill
	\hspace{-0.200cm}
	\begin{minipage}{0.10390\linewidth}
	\centerline{\includegraphics[width=1\linewidth]{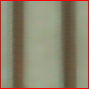}} \vspace{-0.15mm}
	\centerline{\includegraphics[width=1\linewidth]{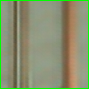}} \vfill \vspace{-0.02cm}
	\centerline{\includegraphics[width=1\linewidth]{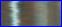}} \vfill \vspace{-0.15cm}
	\centerline{\scriptsize{DPT}}
	\end{minipage}
	\hfill
	\hspace{-0.200cm}
	\begin{minipage}{0.10390\linewidth}
	\centerline{\includegraphics[width=1\linewidth]{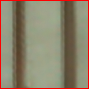}} \vspace{-0.15mm}
	\centerline{\includegraphics[width=1\linewidth]{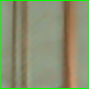}} \vfill \vspace{-0.02cm}
	\centerline{\includegraphics[width=1\linewidth]{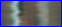}} \vfill \vspace{-0.15cm}
	\centerline{\scriptsize{IINet}}
	\end{minipage}
	\hfill
	\hspace{-0.200cm}
	\begin{minipage}{0.10390\linewidth}
	\centerline{\includegraphics[width=1\linewidth]{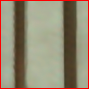}} \vspace{-0.15mm}
	\centerline{\includegraphics[width=1\linewidth]{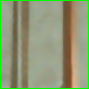}} \vfill \vspace{-0.02cm}
	\centerline{\includegraphics[width=1\linewidth]{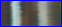}} \vfill \vspace{-0.15cm}
	\centerline{\scriptsize{OFPNet}}
	\end{minipage}
	\end{minipage}
	\end{center}
	\vspace{-7mm}
	\caption{
	  Visual comparisons (central views) of different models (fine-tuned on LytroZoom-O) on the LytroZoom-O testset.
	  Top: $\times 2$ SR.
	  Bottom: $\times 4$ SR.
	  Please zoom in for better visualization and best viewed on the screen.
	  }
	\label{fig:compare2}
	\vspace{-5mm}
\end{figure*}

\begin{figure*}[!t]
	\vspace{-4mm}
	\begin{center}
	\begin{minipage}{0.99\linewidth}
	\begin{minipage}{0.31800\linewidth}
	\vspace{-0.1mm}\centerline{\includegraphics[width=1\linewidth]{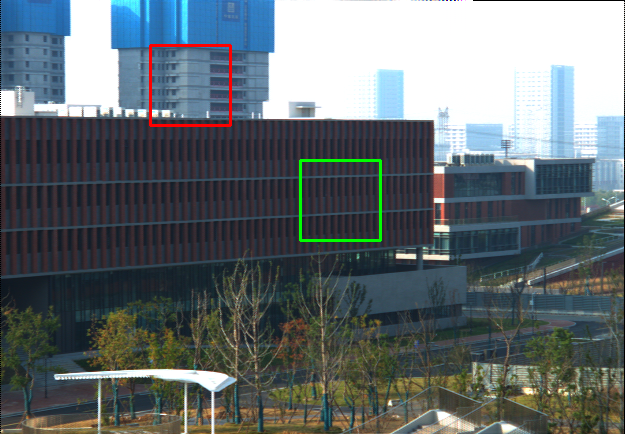}} \vfill \vspace{-0.15cm}
	\centerline{\scriptsize{Outdoor scene captured at 200 mm}}
	\end{minipage}
	\hfill
	\hspace{-0.200cm}
	\begin{minipage}{0.11210\linewidth}
	\vspace{0.5mm}\centerline{\includegraphics[width=1\linewidth]{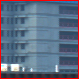}} \vspace{-0.15mm}
	\centerline{\includegraphics[width=1\linewidth]{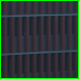}} \vfill \vspace{-0.15cm}
	\centerline{\scriptsize{Input}}
	\end{minipage}
	\hfill
	\hspace{-0.200cm}
	\begin{minipage}{0.11210\linewidth}
	\centerline{\includegraphics[width=1\linewidth]{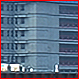}} \vspace{-0.15mm}
	\centerline{\includegraphics[width=1\linewidth]{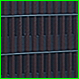}} \vfill \vspace{-0.15cm}
	\centerline{\scriptsize{DPT-BI}}
	\end{minipage}
	\hfill
	\hspace{-0.200cm}
	\begin{minipage}{0.11210\linewidth}
		\centerline{\includegraphics[width=1\linewidth]{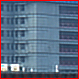}} \vspace{-0.15mm}
		\centerline{\includegraphics[width=1\linewidth]{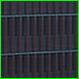}} \vfill \vspace{-0.15cm}
	\centerline{\scriptsize{DPT-LZ-P}}
	\end{minipage}
	\hfill
	\hspace{-0.200cm}
	\begin{minipage}{0.11210\linewidth}
		\centerline{\includegraphics[width=1\linewidth]{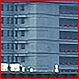}} \vspace{-0.15mm}
		\centerline{\includegraphics[width=1\linewidth]{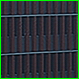}} \vfill \vspace{-0.15cm}
	\centerline{\scriptsize{IINet-BI}}
	\end{minipage}
	\hfill
	\hspace{-0.200cm}
	\begin{minipage}{0.11210\linewidth}
		\centerline{\includegraphics[width=1\linewidth]{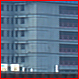}} \vspace{-0.15mm}
		\centerline{\includegraphics[width=1\linewidth]{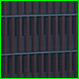}} \vfill \vspace{-0.15cm}
	\centerline{\scriptsize{IINet-LZ-P}}
	\end{minipage}
	\hfill
	\hspace{-0.200cm}
	\begin{minipage}{0.11210\linewidth}
		\centerline{\includegraphics[width=1\linewidth]{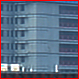}} \vspace{-0.15mm}
		\centerline{\includegraphics[width=1\linewidth]{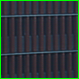}} \vfill \vspace{-0.15cm}
	\centerline{\scriptsize{OFPNet}}
	\end{minipage}
	\end{minipage}
	\end{center}
	\vspace{-7mm}
	\caption{
	  Visual comparisons (central views) of different models on real-world light field scenes ($\times 2$ SR). 
   LZ-P is short for LytroZoom-P.
	  }
	\label{fig:compare3}
	\vspace{-4mm}
\end{figure*}

\vspace{-1mm}
\subsection{Simulated Datasets v.s. LytroZoom}
\vspace{-1mm}
To demonstrate the advantages of the LytroZoom dataset, we conduct experiments to compare the performance of light field SR models trained on simulated datasets and the LytroZoom-P dataset.
We employ the mixed light field datasets~\cite{wang2020spatial} with the angular resolution of $5 \times 5$ to generate simulated $\times 2$ and $\times 4$ light field pairs with the bicubic degradation (BI).
We train InterNet~\cite{wang2020spatial}, IINet~\cite{liu2021intra}, and DPT~\cite{wang2022detail} on BI and LytroZoom-P for each of the two scaling factors ($\times 2$ and $\times 4$).
Since the resolution of the LR-HR pair is the same in LytroZoom-P, we upsample the LR inputs of BI bicubicly and feed the pre-upsampled inputs to the light field SR networks.
We then attach the de-subpixel layer~\cite{vu2018fast} at the beginning of each network to achieve an efficient inference.

We train three networks on BI and LytroZoom-P, respectively, and calculate the average PSNR and SSIM values on the LytroZoom-P testset. 
As shown in Table~\ref{tab:table}, models trained on our LytroZoom-P dataset obtain significantly better performance than those trained on the BI dataset for both scaling factors.
Specifically, the results of the models trained on the BI dataset are even worse than the LR observations.
The reason is apparent: the networks trained with simulated LR-HR light field pairs inevitably get disastrous results when facing complex degradation in real-world scenes.
In Fig.~\ref{fig:bi_lytro}, we visualize the super-resolved central view images obtained by different models.
As can be seen, results generated by models trained on the BI dataset tend to have blurring edges with obvious artifacts.
On the contrary, models trained on LytroZoom-P generate clearer results.

\begin{figure}[!t]
	\begin{center}
	\begin{minipage}{\linewidth}
	\begin{minipage}{0.400\linewidth}
	\vspace{0.5mm} \centerline{\includegraphics[width=1\linewidth,height=0.71\linewidth]{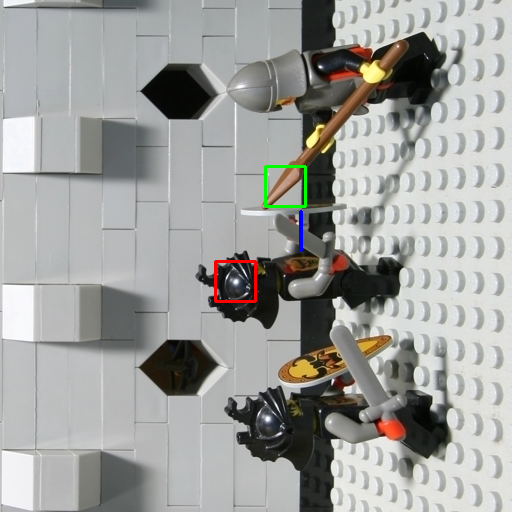}} \vfill \vspace{-0.15cm}
	\centerline{\scriptsize{Lego-Knights}}
	\end{minipage} 
	\hfill
	\hspace{-0.200cm}
	\begin{minipage}{0.11500\linewidth}
		\vspace{0.5mm}\centerline{\includegraphics[width=1\linewidth]{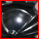}} \vspace{-0.15mm}
	\centerline{\includegraphics[width=1\linewidth]{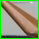}} \vfill \vspace{-0.02cm}
	\centerline{\includegraphics[width=1\linewidth,height=0.45\linewidth]{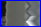}} \vfill \vspace{-0.15cm}
	\centerline{\scriptsize{Input}}
	\end{minipage}
	\hfill
	\hspace{-0.200cm}
	\begin{minipage}{0.11500\linewidth}
	\centerline{\includegraphics[width=1\linewidth]{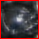}} \vspace{-0.15mm}
	\centerline{\includegraphics[width=1\linewidth]{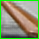}} \vfill \vspace{-0.02cm}
	\centerline{\includegraphics[width=1\linewidth,height=0.45\linewidth]{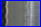}} \vfill \vspace{-0.15cm}
	\centerline{\scriptsize{InterNet}}
	\end{minipage}
	\hfill
	\hspace{-0.200cm}
	\begin{minipage}{0.11500\linewidth}
	\centerline{\includegraphics[width=1\linewidth]{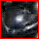}} \vspace{-0.15mm}
	\centerline{\includegraphics[width=1\linewidth]{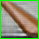}} \vfill \vspace{-0.02cm}
	\centerline{\includegraphics[width=1\linewidth,height=0.45\linewidth]{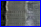}} \vfill \vspace{-0.15cm}
	\centerline{\scriptsize{DPT}}
	\end{minipage}
	\hfill
	\hspace{-0.200cm}
	\begin{minipage}{0.11500\linewidth}
	\centerline{\includegraphics[width=1\linewidth]{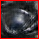}} \vspace{-0.15mm}
	\centerline{\includegraphics[width=1\linewidth]{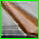}} \vfill \vspace{-0.02cm}
	\centerline{\includegraphics[width=1\linewidth,height=0.45\linewidth]{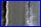}} \vfill \vspace{-0.15cm}
	\centerline{\scriptsize{IINet}}
	\end{minipage}
	\hfill
	\hspace{-0.200cm}
	\begin{minipage}{0.11500\linewidth}
	\centerline{\includegraphics[width=1\linewidth]{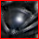}} \vspace{-0.15mm}
	\centerline{\includegraphics[width=1\linewidth]{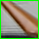}} \vfill \vspace{-0.02cm}
	\centerline{\includegraphics[width=1\linewidth,height=0.45\linewidth]{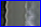}} \vfill \vspace{-0.15cm}
	\centerline{\scriptsize{OFPNet}}
	\end{minipage}
	\end{minipage}
	\end{center}
	\vspace{-7mm}
	\caption{
	  Visual comparisons (central views) of different models on real-world light field scene captured by Gantry.
	  }
	\label{fig:compare4}
	\vspace{-2mm}
\end{figure}

\vspace{-1mm}
\subsection{Baseline Methods v.s. OFPNet}
\vspace{-1mm}
We compare our proposed OFPNet with three selected baseline methods.
All models are trained/fine-tuned and tested on LytroZoom-P and LytroZoom-O.

\noindent\textbf{LytroZoom-P.}
As is shown in Table~\ref{tab:table}, our OFPNet earns the highest PSNR and SSIM values.
For example, one can see that OFPNet surpasses IINet~\cite{liu2021intra}, the state-of-the-art light field SR method, by 0.29dB/0.0080 on $\times 4$ SR in terms of PSNR/SSIM.
Fig.~\ref{fig:compare} shows the super-resolved central view images for qualitative comparison.
In terms of visual quality, OFPNet beats previous methods, providing fine details without introducing unappealing artifacts in general.
For example, as seen in Fig.~\ref{fig:compare}, baseline methods can hardly restore the external shape of the buildings in \textit{Bangkok2}.
In contrast, our proposed OFPNet can generate results with vivid details and patterns.

\noindent\textbf{LytroZoom-O.}
As is shown in Table~\ref{tab:table2}, on the LytroZoom-O testset, our OFPNet also earns the highest PSNR and SSIM values.
Fig.~\ref{fig:compare2} shows the super-resolved central view images from LytroZoom-O for qualitative comparison.
OFPNet can generate results with fine details.

\begin{table}[!t]
\caption{
	Analysis of the OFPNet on $\times 4$ SR on LytroZoom-P.
}
\centering
\vspace{-3mm}
\renewcommand\arraystretch{1.1}
\resizebox{0.88\linewidth}{!}{
\begin{tabular}{|p{1cm}<{\centering} | p{1cm}<{\centering} | p{1cm}<{\centering}|c|c|c|}
	\hline
	\multicolumn{3}{|c|}{Frequency decomposition} & \multicolumn{2}{c|}{Frequency Projection} & \multirow{2}{*}{PSNR} \\ \cline{1-5}
	$\mathcal{F}_l$ &$\mathcal{F}_m$ &$\mathcal{F}_h$ & Interactions        & FP operation        &                       \\ 
	\hline
	\hline
	\scriptsize\XSolidBrush & \scriptsize\XSolidBrush  & \checkmark & -                & -                 & 29.87               \\ \hline
	\scriptsize\XSolidBrush & \checkmark  &\checkmark  & -                & -                 & 29.98           \\ \hline
	\checkmark & \checkmark  &\checkmark  & -                & -                 & 30.11                   \\ 
	\hline
	\hline
	- & -  &-  & \scriptsize\XSolidBrush &\scriptsize\XSolidBrush  & 29.60           \\ \hline
	- & -  &-  & \checkmark &\scriptsize\XSolidBrush  & 29.86           \\ \hline
	- & -  &-  & \scriptsize\XSolidBrush &\checkmark  & 29.99          \\ \hline
	- & -  &-  & \checkmark &\checkmark  & 30.11    \\ 
	\hline
	\end{tabular}  
}
\vspace{-4mm}
\label{tab:ablation}
\end{table}

\vspace{-1mm}
\subsection{Generalization Tests}
\vspace{-1mm}
Our LytroZoom-trained light field SR models exhibit robust generalization capabilities, both in terms of content and device.
Specifically, our LytroZoom-P-trained models perform well on a scene captured at 200 mm focal length, as demonstrated in Fig.~\ref{fig:compare3}, despite being trained on indoor scenes printed on postcards.
For the device generalization, as shown in Fig.~\ref{fig:compare4}, the light field SR models trained on the dataset captured by a Lytro ILLUM camera can be readily applied to different devices such as Gantry (we super-resolve the input light field directly from the scene \textit{Lego-Knights} in STFgantry~\cite{StfGantry}).

\vspace{-2mm}
\subsection{Model Analysis}
\vspace{-2mm}

\noindent \textbf{Investigation of the frequency decomposition.}
We investigate different extracted frequency components in OFPNet.
We have added several residual blocks after the extracted frequency features while removing the corresponding components in Table~\ref{tab:ablation} to ensure the parameters remain unchanged.
As shown in Table~\ref{tab:ablation}, our results demonstrate that incorporating additional frequency components in OFPNet leads to improved performance in terms of PSNR, with gains of 0.24 dB and 0.13 dB when considering higher frequency components.
These results suggest that the omni-frequency components play a crucial role in real-world light field SR.
Further details and analysis can be found in the supplementary document.

\noindent \textbf{Investigation of frequency projection.}
When we simultaneously remove the interactions between frequency components and the FP operations (we replace the FP operations with residual blocks of the same parameters), we only get a result of 29.60dB in terms of PSNR.
When we add interaction and FP operations, the results improve by 0.26dB and 0.39dB, respectively, indicating the importance of these two designs.
The best result can be obtained using these two designs simultaneously (30.11dB) on LytroZoom-P.

\section{Discussion}
\vspace{-2mm}
Despite the encouraging performance as shown above, the LytroZoom dataset still has certain limitations.
(1) The use of a single data collection device, specifically a Lytro ILLUM camera, limits the generalizability of models trained on LytroZoom to other cameras. While these models can perform well on the Gantry dataset, there may still be a domain shift when applied to light fields captured by cameras with different baselines, resulting in artifacts (as seen in Fig.~\ref{fig:compare4}). To address this limitation, future work can expand the LytroZoom dataset to include light fields captured by other types of cameras, such as camera arrays and Gantry.
(2) Minor distortions that cannot be rectified.
The registration step~\cite{cai2019toward} could alleviate the distortions caused by different FoVs, yet minor misalignment and luminance/color differences exist between the LR-HR light fields.
It is likely due to these minor distortions that cause the non-convergence of ATO~\cite{jin2020light} and DistgSSR~\cite{wang2022disentangling} during the training stage~\cite{yang2021real}.
We will investigate new training strategies.
(3) Inadequate benchmark experiments.
We will extend LytroZoom with more scaling factors and conduct experiments on scale-arbitrary real-world light field SR.

\vspace{-2mm}
\section{Conclusion} 
\vspace{-2mm}
In this paper, we have collected the first real-world paired LR-HR light field SR dataset, \textit{i.e.}, LytroZoom, with authentic degradation for real-world light field SR.
Specifically, a Lytro ILLUM camera is used to collect accurate pixel-wise aligned LR and HR light field pairs on 94 city scenes printed on postcards (LytroZoom-P) and 63 outdoor static objects (LytroZoom-O).
A novel OFPNet is proposed to efficiently solve the real-world light field SR problem by considering omni-frequency information and enhancing cross-frequency representations.
Our extensive experiments validate that the models trained on our LytroZoom dataset can lead to much better real-world light field SR results than trained on simulated datasets.
They are also readily generalizable to diverse content and devices.
We believe that LytroZoom presents a valuable opportunity for the research community to further explore real-world light field SR.

\noindent \textbf{Acknowledgement.}
This work was supported in part by the National Natural Science Foundation of China under Grants 62131003 and 62021001.


{\small
\bibliographystyle{ieee}
\bibliography{PaperForReview}
}

\end{document}

%% file: includesAndMacros_yagiz.tex
\usepackage{graphics}
\usepackage[usenames,dvipsnames,table]{xcolor}
\usepackage{array}
\usepackage{multirow}
\usepackage{wrapfig}
\usepackage{hhline}
\usepackage{pifont}
\newcolumntype{K}[1]{>{\centering\arraybackslash}p{#1}}
